\documentclass[acmtog,screen,nonacm]{acmart}

\AtBeginDocument{%
  }

\graphicspath{{figures/}}
\usepackage{subcaption}
\usepackage{csquotes}
\usepackage{cleveref}

\crefname{figure}{Fig.}{Figs.}
\Crefname{figure}{Fig.}{Figs.}
\crefname{equation}{Eq.}{Eqs.}
\Crefname{equation}{Eq.}{Eqs.}
\crefname{section}{Sec.}{Secs.}
\Crefname{section}{Sec.}{Secs.}
\crefname{table}{Tab.}{Tabs.}
\Crefname{table}{Tab.}{Tabs.}

\newcommand{\deepsdf}{DeepSDF}

\newcommand{\etal}{\textit{et al.}}

\newcommand{\model}{f} 

\newcommand{\point}{\mathbf{p}} 
\newcommand{\cpoint}{\mathbf{c}} 
\newcommand{\latent}{\mathbf{z}} 
\newcommand{\sd}{s} 
\newcommand{\dsd}{\Delta s} 
\newcommand{\weight}{w} 

\newcommand{\dpoint}{\Delta \point} 
\newcommand{\sindicator}{\delta} 

\newcommand{\spatial}{\Omega} 
\newcommand{\surface}{\mathcal{S}} 

\newcommand{\lbl}{\ell} 

\newcommand{\rnet}{\mathcal{R}} 
\newcommand{\dnet}{\Phi} 
\newcommand{\hnet}{\Psi} 

\newcommand{\loss}{\mathcal{L}} 
\newcommand{\centroid}{\mathrm{centroid}} 
\newcommand{\segmentation}{\mathrm{seg}} 

\DeclareMathOperator{\bce}{BCE} 

\begin{document}

\title{An Implicit Parametric Morphable Dental Model}

\author{Congyi Zhang}
\affiliation{%
  \institution{The University of Hong Kong}
  \country{Hong Kong}}
\affiliation{%
  \institution{Max Planck Institute for Informatics}
  \city{Saarbrücken}
  \country{Germany}}
\email{cyzhang@cs.hku.hk}

\author{Mohamed Elgharib}
\affiliation{%
  \institution{Max Planck Institute for Informatics}
  \city{Saarbrücken}
  \country{Germany}}
\email{elgharib@mpi-inf.mpg.de}

\author{Gereon Fox}
\affiliation{%
  \institution{Max Planck Institute for Informatics}
  \city{Saarbrücken}
  \country{Germany}
}
\email{gfox@mpi-inf.mpg.de}

\author{Min Gu}
\affiliation{%
 \institution{The University of Hong Kong}
 \country{Hong Kong}}
\email{drgumin@hku.hk}

\author{Christian Theobalt}
\affiliation{%
  \institution{Max Planck Institute for Informatics}
  \city{Saarbrücken}
  \country{Germany}}
\email{theobalt@mpi-inf.mpg.de}

\author{Wenping Wang}
\affiliation{%
  \institution{Texas A\&M University}
  \city{College Station}
  \country{USA}}
\affiliation{%
  \institution{The University of Hong Kong}
  \country{Hong Kong}}
\email{wenping@cs.hku.hk}

\renewcommand{\shortauthors}{Zhang et al.}

\begin{abstract}

3D Morphable models of the human body capture variations among subjects and are useful in reconstruction and editing applications. Current dental models use an explicit mesh scene representation and model only the teeth, ignoring the gum. In this work, we present the first parametric 3D morphable dental model for both teeth and gum. Our model uses an implicit scene representation and is learned from rigidly aligned scans. It is based on a component-wise representation for each tooth and the gum, together with a learnable latent code for each of such components. It also learns a template shape thus enabling several applications such as segmentation, interpolation and tooth replacement.
Our reconstruction quality is on par with the most advanced global implicit representations while enabling novel applications. 
Project page: \url{https://vcai.mpi-inf.mpg.de/projects/DMM/}

\end{abstract}

\ccsdesc[300]{Computing methodologies~Shape modeling}
\ccsdesc[300]{Computing methodologies~Shape representations}

\keywords{Implicit neural representation, Morphable model, Teeth}

\begin{teaserfigure}
  \includegraphics[width=\textwidth]{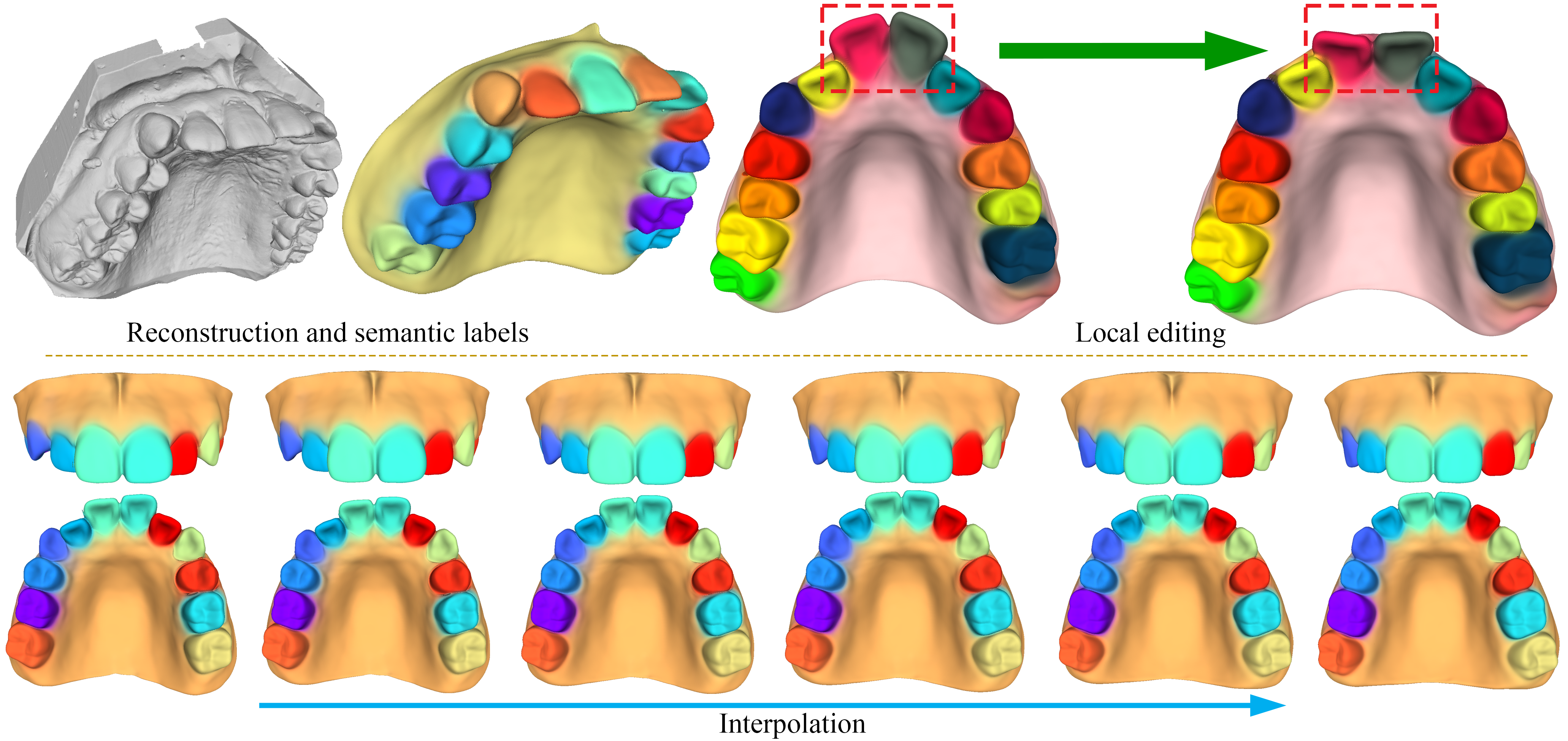}
  \caption{
  We present an SDF-based morphable model for the human teeth and gums. Our model is \emph{compositional}, i.e. the full geometry is a combination of a number of smaller components, that each model one semantically meaningful component: Each tooth and also the gums are controlled by separate latent codes. This allows our model to not just reconstruct geometry, but to also compute a semantic labelling in addition. Furthermore, it enables editing of specific components, e.g. individual teeth (see dashed red). Our model can also be used to smoothly interpolate between different teeth configurations, possibly serving as a visual aid in the communication between orthodontists and their patients.
  }
  \label{fig:teaser}
\end{teaserfigure}
\maketitle

\section{Introduction}

The availability of morphable human face models has enabled various applications such as 
virtual face avatars for telecommunication \cite{Lombardi:2018,Wang2021FaceVid2Vid},
photorealistic animation in movies and media production~\cite{Flawless,Synthesia,Kim2019}, 
single-photo editing~\cite{Tewari2020PIE}, and others~\cite{Zollhoefer2018FaceSTAR,Thies19DNR,Hu17}. So far, however, most of such methods omit the modelling of the mouth interior, in particular teeth and gum.
It is not only in the aforementioned applications that these parts of the human face are of importance, but also in medical research, for instance in orthodontic treatment. 
Capturing the geometry of the dental region (teeth and gum) is the basis for many interesting use cases, such as planning of the treatment or visualizing expected results for the patient. In this context, the availability of a morphable model with some control over original geometric components, e.g. over single tooth, could for example allow animating a transition from the status quo in a patient towards a desired treatment result.

Processing the human teeth poses multiple challenges: For one, human teeth can be shifted, rotated and generally misaligned in many ways, and some of them may even be missing altogether. Furthermore, the topological variety of teeth and their very uniform texture make it very hard to reliable detect, or even define any stable features on them. While there have been attempts to create morphable models for the human teeth only~\cite{wu2016teeth} (without gum), they are based on explicit representations such as meshes with a manually defined template shape. Learning such models requires accurate non-rigid registration of 3D scans and manual labeling of the teeth during reconstruction. Until now, there is no parametric morphable model for the human teeth that also includes the gums.

In this work, we present the first parametric morphable dental model for the geometry of the human teeth and gums. While the vast majority of human body models use explicit representations such as meshes~\cite{SMPL:2015,Romero2017MANO,3DMM_survey}, recently, there has been strong interest in using implicit representations~\cite{Yenamandra2021i3DMM,Zheng2022ImFace,Corona2022LISA,Palafox2021NPMs,Alldieck21}.
Motivated by this, we use, for the first time, an implicit representation for modeling  teeth and gums:
We adapt the implicit representation of DeepSDF~\cite{Park19DeepSDF} where an object's geometry is represented by the decision boundary of a classifier that is supposed to tell whether points lie inside or outside of the examined object. 
To further identify dense correspondences, we learn a template shape and its deformations with the help of Hyper Networks~\cite{Sitzmann2020Siren}. 
We use a compositional DeepSDF representation, i.e. each tooth or the gum is assigned a separate DeepSDF network with a learnable latent code.
In addition to improving reconstruction accuracy over most of the original \emph{global} implicit representations~\cite{Park19DeepSDF,Zheng2021DIT}, the compositional representation allows various editing applications such as tooth replacement and interpolation (see Fig.~\ref{fig:teaser}). 
Our model can be learned from merely aligned 3D scans.
The final overall model is a combination of the outputs of all the component models, weighted by segmentation indicators that are also predicted by the model.

In summary, we make the following contributions:
\begin{itemize}
    \item We present the first implicit morphable model for the geometry of the human teeth and gum. We use a compositional implicit representation, with learnable latent codes for each region.
    \item Our model learns a template shape, which automatically establishes correspondences that help predict geometry segmentation during reconstruction. 
    \item We introduce novel segmentation and teeth centroid losses that are crucial for training the model.
    \item Our technique produces accurate reconstructions that are on par with the state of the art, in addition to enabling novel applications such as tooth replacement, interpolation and segmentation.
\end{itemize}

As of yet, there is no publicly accessible model of the human teeth and gums, making ours the first such model when we release it.
\section{Related Work}

In this section we start by discussing implicit representations used for modeling the scene geometry. Here, we examine both global~\cite{Park19DeepSDF,Mescheder19OccNet,Chen2019ImplicitDecoder} and local~\cite{Zheng2021DIT,Deng2021DIF} representations, and the advantages the latter brings to literature. We then discuss recent methods for modelling the human body using implicit representations. Finally, we discuss methods for teeth processing with emphasize on teeth reconstruction and modeling.

\subsection{Scene geometry modeling}

Modeling scene geometry is a fundamental task in both computer vision and computer graphics. Previous methods use explicit representations such as meshes~\cite{Thies18Face2Face}, voxels~\cite{Niessner2013Hashing} or points clouds~\cite{Keller2013}. While such explicit representations have been successful in many applications~\cite{Zollhoefer2018RecoSTAR}, they suffer from a number of limitations: Voxels are memory-intensive, meshes struggle to handle detailed structures and point clouds are sparse and lose a significant portion of the geometry. Thus, in the past few years several efforts were made to explore so-called \emph{implicit} geometrical representation~\cite{Park19DeepSDF,Mescheder19OccNet,Chen2019ImplicitDecoder}. Unlike the earlier approaches, implicit representations encode the geometry indirectly, for instance as the decision boundary of a classifier that decides whether a point lies inside or outside the examined object. Among the most popular implicit representations are DeepSDF~\cite{Park19DeepSDF} and Occupancy Networks~\cite{Mescheder19OccNet}. DeepSDF uses a signed distance function to measure how far a point is from the surface of the examined object, while Occupancy Networks predict the probability of a point lying inside the object. Both methods demonstrate interesting capabilities such as interpolating between latent codes learned either by an auto-decoder~\cite{Park19DeepSDF} or an auto-encoder~\cite{Mescheder19OccNet} architecture.

Follow-up works addressed limitations of implicit representations. One main limitation is the lack of correspondences, which limit editing and model learning capabilities. To this end, some works proposed to learn a template that is shared by all the training samples, which may be similar to the mean shape for the training samples. For instance, \enquote{Deep Implicit Templates}\cite{Zheng2021DIT}, or DIT for short, uses a network that learns the deformation to the template shape.
\enquote{Deformed Implicit Field} \cite{Deng2021DIF} , or DIF, proposes a similar idea, but, inspired by~\cite{Sitzmann2019SRNs}, they use so-called Hyper-Nets, that predict the weights of their deformation networks.

There are several implicit representations that decompose the geometry into a number of localized components~\cite{Wu2020PQNET,Genova2020LDIF,Peng2020ECCV,Chabra2020,Tretschk2020PatchNets,Chen2021Multiresolution} as opposed to the single global representation used in the earlier works~\cite{Park19DeepSDF,Mescheder19OccNet,Chen2019ImplicitDecoder}. The motivation here is that while implicit representations are powerful, their global formulation could limit their generalization capabilities and reconstruction accuracy. Genova~\etal~\cite{Genova2020LDIF} use localized deep implicit representations and assign latent codes to each local region. The method estimates a template shape as well as geometrical details through a local shape encoder. However, it automatically decomposes the watertight shape into a pre-defined number of components, the definition of which cannot be controlled and would lead to problems and mis-shaped geometry when teeth are missing. Chabra~\etal~\cite{Chabra2020} uses a localized SDF representation with local latent codes defined in a voxel grid. Both methods show finer geometric details than global formulations and better generalization capabilities. Another potentially useful application of localized representations is the ability to perform novel applications. For instance Yin~\etal~\cite{Yin20} proposed a method that combines different parts of the same object together. This is done by learning the connections/joints of the various parts. Joints are learned using  the implicit representation of Chen~\etal~\cite{Chen2019ImplicitDecoder}. They are learned in a way to agree with the remaining components while being smooth and topologically valid. The solution is trained with segmented components extracted from ShapeNet. PQ-Net \cite{Wu2020PQNET} represents and generates 3D shapes in a sequential part assembly manner. However, since this method is not able to automatically segment the input data, it requires segmentation annotations at test time for the reconstruction task. Their geometric components are rigidly assembled to compose a model, while we believe that dental models require smooth blending of components into one reconstruction.

\subsection{Implicit-based Modeling}
\label{sec:implicitmodeling}
Recently there has been increasing interest in building models of the various parts of the human body using implicit representations. This includes models for the human head~\cite{Yenamandra2021i3DMM,Zheng2022ImFace}, hands~\cite{Corona2022LISA} and body~\cite{Deng2020NASA,Palafox2021NPMs,Alldieck21}. 
The work of Yenamandra~\etal~\cite{Yenamandra2021i3DMM} was the first in this regard. They presented the first 3D morphable model of the human head, including hair. The model learns latent codes for identity, albedo, expression and hairstyle. The model, named i3DMM, is based on a SDF-based architecture learned from 3D scans of various subjects with different hairstyles and performing different expressions. The method learns a template shape and a deformation to this shape. The template shape establishes correspondences and hence, unlike early 3DMM explicit face models~\cite{3DMM_survey}, it does not need complicated non-rigid alignment of the scans, but merely rigidly aligned ones. The method shows novel interpolation applications in the latent spaces of all components e.g. identity, expressions and hairstyle.
ImFace~\cite{Zheng2022ImFace} is a concurrent work to the one we present here. The aim is to improve the reconstruction accuracy of i3DMM~\cite{Yenamandra2021i3DMM} using a localized SDF representation. To this regards, the entire face is decomposed into 5 regions with separate networks for expression and identity learning. A meta-learning approach is used, where hyper-nets learn the weights of the expression and identity networks. Results show more accurate reconstructions over i3DMM~\cite{Yenamandra2021i3DMM}.
An important difference between ImFace and our work is that since we aim at providing semantic control over each tooth individually, we assign one dedicated latent code to each geometric component, i.e. to each tooth and to the gums, whereas ImFace's latent codes cannot be partitioned into distinct regions of the geometry.
Also, our method is different from NASA \cite{Deng2020NASA}: For a fixed number of joints, NASA encodes geometry as pose-conditioned occupancy. This is ill-suited for dental geometry, as single teeth might be missing, and annotating individual teeth poses and skinning weights for the dental scans in the training set would be very  difficult.

\subsection{Human Teeth Processing}

There are several methods for processing the human teeth, including methods for teeth reconstruction~\cite{Zheng11,Farag13,Abdelrehim2014PCA,Wirtz2021Teethrecon,wu2016teeth}, restoration and completion~\cite{Mostafa2014CSM,Ping2021Completion}, orthodontic treatment~\cite{Yang2020iOrthoPredictor}, segmentation~\cite{Cui21,Zhang_2021_CVPR}, pose estimation~\cite{Murugesan18,Beeler14,yang2019building} and others~\cite{Velinov18,Wei20}.  The closest to our work are methods for teeth reconstruction and restoration~\cite{Abdelrehim2014PCA,Mostafa2014CSM,wu2016teeth,Wirtz2021Teethrecon,Ping2021Completion}. The vast majority of these methods use explicit representations, with the exception of Ping~\etal~\cite{Ping2021Completion}. This work, however, does not propose a morphable model,  but rather takes teeth crowns as input and completes them with gum. 
Abderlrehim~\etal~\cite{Abdelrehim2014PCA} reconstructs individual teeth from a single image by using shape-from-shading with shape priors. Here, shape priors are learned using PCA on height maps. 
Mostafa~\etal~\cite{Mostafa2014CSM} proposes a method for tooth restoration based on a single captured image. Restoration is achieved by aligning to a model derived from ensemble of oral cavity shapes and textures.
Wirtz~\etal~\cite{Wirtz2021Teethrecon} reconstructs teeth only, without gums, from 5 images of different viewpoints.
They propose a model-based method that deforms a mean teeth shape to fit the input images.
The optimum fitting is estimated by minimizing a 
silhouette loss between the 2D projections of the fitted 3D model and the observed 2D images.

Wu~\etal~\cite{wu2016teeth} utilizes a teeth model learned from high quality dental scans and fit it to multiple images shot from either a multiview camera setup or from a handheld moving camera.
The teeth model follows a mesh representation and does not include the gum. To build the model, the 3D scans are aligned with respect to a manually defined template. Alignment is done through user-intervention, followed by a combination of rigid and non-rigid registration. An average shape model is estimated and PCA is used to estimate local shape variations. The final model accounts for the global deformations of the entire tooth row and the variations of each tooth individually.

We present the first implicit-based model for the human teeth and gum. It is morphable by design, producing intermediate shapes by interpolating between latent codes. Furthermore, it learns a reference shape that allows the computation of correspondences. This produces segmentation masks of the teeth as a by-product, as well as allowing interesting editing applications such as teeth replacement. In comparison to Wu~\etal~\cite{wu2016teeth}, our model does not need any labeling of the teeth for reconstruction, except for a binary vector indicating the presence/absence of individual teeth. Our model is trained on scans that were aligned only rigidly, in contrast to Wu~\etal, who require sophisticated non-rigid alignment coupled with user-intervention. It also jointly models the gum with the teeth, unlike Wu~\etal, who only model teeth. We believe our model is a useful contribution to recent efforts of building implicit models of the full human body as discussed in \cref{sec:implicitmodeling}. We will release our model for research purposes, thus making it the only publicly available morphable teeth model including gums.

\section{Overview}\label{sec:overview}

\begin{figure}[htbp]
\includegraphics[width=0.45\columnwidth]{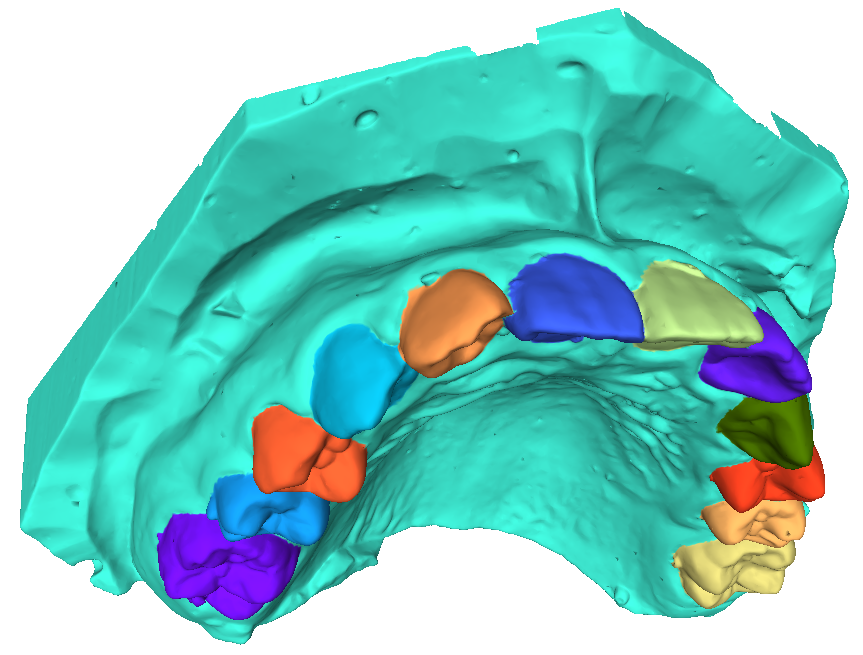}
\includegraphics[width=0.45\columnwidth]{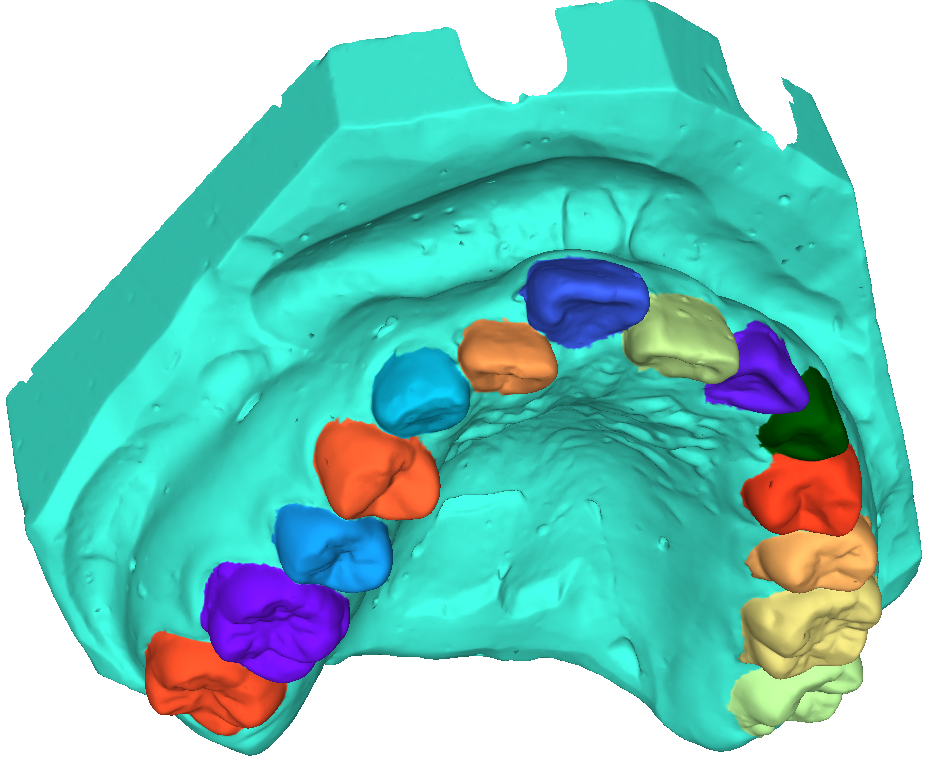}
\caption{Examples from our dataset of ground truth teeth geometries. Teeth identities have been annotated manually. We visualize them by different colours.
}
\label{fig:example_of_raw_data}
\end{figure}

\begin{figure*}[t]
\subcaptionbox[]{Overall workflow\label{subfig:overall_net}}{\includegraphics[width=0.9\textwidth]{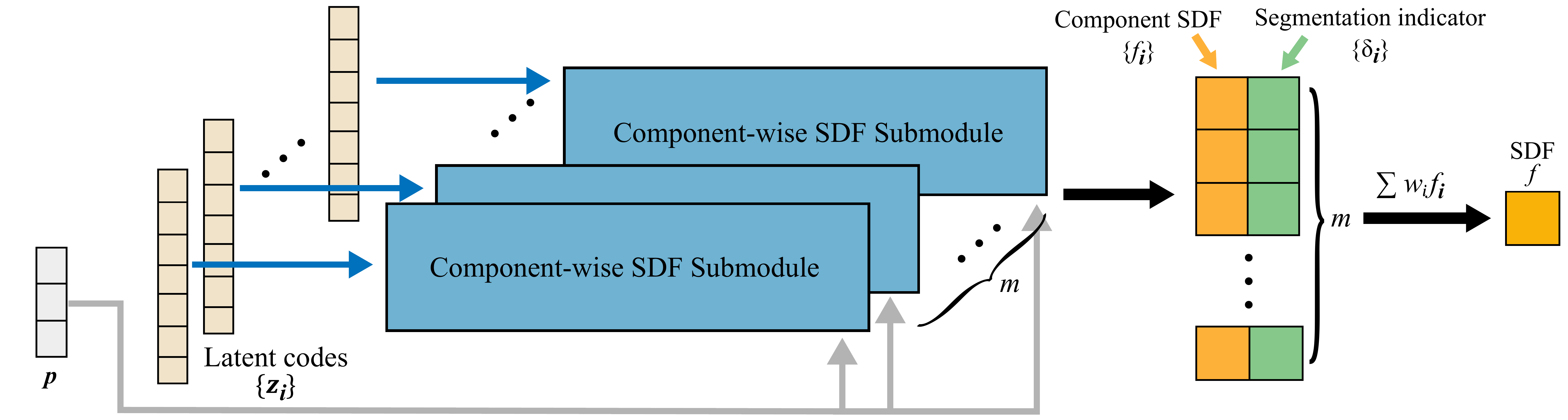}}\\
\subcaptionbox[]{Sub-network for one component\label{subfig:component_net}}{\includegraphics[width=0.9\textwidth]{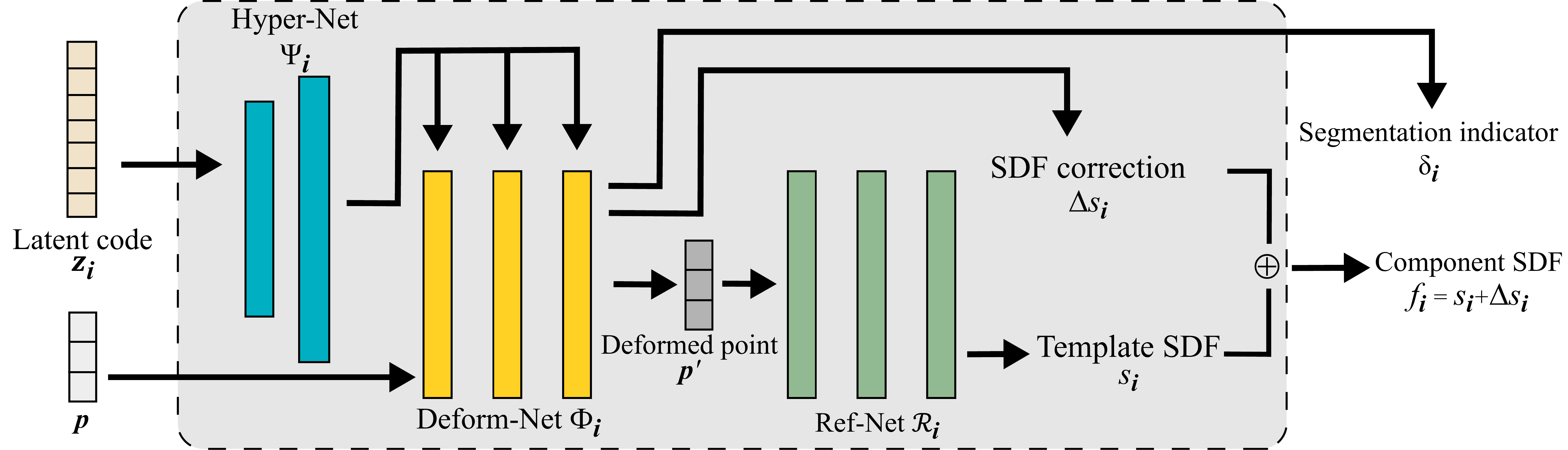}}
  \caption{The pipeline of our proposed method. We use a component-wise SDF representation where each tooth and the gum is represented by a separate "Component Shape Model". These models learn a reference shape for each component (in the Ref-Net), that is queried at those points to which the input points are warped by Deform-Net.
  Based on the component-wise SDF values and the segmentation indicators $\sindicator$ predicted for each component, we compute the full geometry as a weighted sum (see top right).
}
\label{fig:overview}
\end{figure*}

Our aim is to build a morphable model for the geometry of the human teeth and gum, with the ability to control each component individually. To this end, we present a compositional SDF representation where separate models are used to represent each tooth and the gum. For a standard dental scan (maxilla or mandible), we typically have one gum and up to 14 teeth (excluding 2 wisdom teeth), and thus we build our model assuming $m=15$ components in total (\cref{fig:example_of_raw_data}).
We learn 1 dedicated latent code for each of these 15 components. This allows editing applications such as tooth replacement and morphing.

Our proposed network consists of $m$ sub-modules that represent the different components of the dental scan (\cref{subfig:overall_net}). Conditioned on a latent code and given a point in 3D space as input, each sub-module $i$ predicts an SDF value $\sd_i + \dsd_i$ and an indicator $\sindicator_i$. The latter provides an estimate of the probability that the input point belongs to that part of the geometry that the sub-module is responsible for.
Based on the $\sindicator_i$ values for all components, we compute a set of blending weights $\weight_i$ to linearly combine the accompanying SDF values to one final value.

For each sub-module, we use a variant of DIF \cite{Deng2021DIF} (as shown in \cref{subfig:component_net}): The input spatial point will first be warped by a Deform-Net. This network is generated by a Hyper-Net that is conditioned on a latent code. 
It maps each input point to a learned canonical reference space, in which a template shape is embedded by a Ref-Net (see \cref{subfig:component_net}). Deform-Net also predicts the SDF compensation $\dsd_i$ to refine geometric details. Hence the component-wise SDF is computed as $\sd_i + \dsd_i$, where $\sd_i$ is the SDF value predicted by the Ref-Net.

Our method is trained on a dataset of dental geometries, manually annotated with semantic labels that segment the surface into the individual tooth types and the gums (see Fig.~\ref{fig:example_of_raw_data}). Even though these models have been acquired by different methods (e.g. by digitizing the traditional teeth impression of a patient, or more directly by an intra-oral scanning method) we will refer to them mostly as \enquote{dental scans}, to avoid confusion between the various meanings of the word \enquote{model}. Each scan is available as a high-resolution mesh, in particular allowing us to obtain normal vectors for supervision. More details are given in \cref{sec:implementation}

\section{Method}

To enable control over each geometric component individually, we decompose the overall latent space of the entire model into sub-spaces that correspond to teeth and gums respectively.
Inspired by \deepsdf{}~\cite{Park19DeepSDF}, we model the entirety of the teeth scan geometry as a function $\model$ that is conditioned on a set $\{\latent_i\}_{i=1,\dots,m}$ of gum and teeth latent codes (where $m=15$ is the number of geometric components in our implementation), and maps arbitrary spatial locations $\point$ to the signed distance $\sd$ between $\point$ and the surface:

\begin{equation}
\model(\point, \latent_1,\ldots,\latent_m) = \sd
\end{equation}

where the set  $\{ \point \mid \model(\point, \latent_1,\dots,\latent_m)=0\}$ constitutes the surface of the model.
To learn sub-spaces of the geometry of individual components, we decompose the overall function $\model$ into a set $\{\model_i\}_{i=1,\dots,m}$ of sub-functions:

\begin{equation}\label{eq:linear_combination}
    \model(\point, \latent_1,\ldots,\latent_m) = \sum_{i=1}^m \weight_i \cdot \model_i(\point, \latent_i) = \sum_{i=1}^m \weight_i \cdot (\sd_i + \dsd_i)
\end{equation}

where $\weight_i$ are blending weights $\sd_i$ are signed distance values for component $i$ and $\dsd_i$ are small-scale correction offsets for these values.
In the following sections, we discuss how to compute the blending weights $\weight_i$ and the function values $\model_i(\point, \latent_i) = \sd_i + \dsd_i$, and we will define the objective function that we use for supervision.

\subsection{Network structure}
Each sub-function $\model_i$ is implemented as a pair $\dnet_i, \rnet_i$ of neural networks (see \cref{subfig:component_net}): The \emph{reference shape network} $\rnet_i(\point^\prime_i) = \sd_i$ predicts SDF values $\sd_i$ for the $i$-th reference shape, independent of the latent code $\latent_i$.
It is queried at those points $\point^\prime_i$ that are produced by the \emph{deformation network} $\dnet_i(\point, \latent_i) = (\point^\prime_i, \dsd_i, \sindicator_i)$.
The \emph{deformation network} is supposed to warp the space in which the reference shape is embedded, such that querying $\rnet_i$ at the points $\point^\prime_i$ results in the shape encoded by the $\latent_i$.

In terms of output, there is a slight difference between the deformation networks for the gums and those for the teeth:  For the gums,  deformation is represented directly by the offset $\dpoint$ of each query point $\point$ so that $\point^\prime=\point+\dpoint$. For the teeth however, the points close to them are always subject to almost the same rotation and translation, because teeth have characteristic rigid shapes. Therefore, inspired by \cite{Park2021Nerfies}, we represent the transform for each point by a screw axis $(\mathbf{r};\mathbf{t})\in\mathbb{R}^6$ and set $\point^\prime=e^\mathbf{r}\point+\mathbf{t}$ by Rodrigues' formula \cite{rodrigues1816}.

Note that the weights of $\dnet_i$ are themselves the output of a so-called Hyper-Net $\hnet_i(\latent_i)$, a mechanism introduced in recent works \cite{Sitzmann2020Siren, Deng2021DIF}.
We differ from previous uses, however, in that $\dnet_i$ not only predicts the deformed points $\point^\prime_i$ and SDF value correction deltas $\dsd_i$, but also an additional value $\sindicator_i\in[0,1]$ (see \cref{subfig:component_net}), which serves as an indicator for teeth segmentation. This indicator is used to estimate the blending weights from \cref{eq:linear_combination}:
\[\weight_i := \frac{\delta_i}{\sum_{j=1}^m\delta_j}\]

\subsection{Objective Function}
\label{sec:method}

To train our component-wise implicit neural representations, we not only build on loss terms from previous work \cite{Deng2021DIF, Sitzmann2020Siren}, but also contribute  two new loss terms that help decompose the geometry into semantically meaningful components, namely our \emph{centroid loss} and our \emph{segmentation loss}.

\paragraph{Centroid loss.}
To guide and regularize the deformation field, most face-related works use landmarks as their spatial constraints for deformation \cite{Yenamandra2021i3DMM,Zheng2022ImFace}. However, it is not easy to detect or even define landmarks on the teeth geometry.
We solve this problem by enforcing that deforming the centroid point $\cpoint_i$ of tooth $i$ leads to a result $\cpoint^\prime_i$, that should coincide with the average centroid position $\bar{\cpoint}_i$ of the training data for that tooth, which we can precompute before training.
We penalize the $\ell^1$ distance between the predicted and the expected centroid:
\begin{equation}
\loss^\centroid_i=\|\cpoint^\prime_i - \bar{\cpoint}_i\|_1
\end{equation}

\paragraph{Segmentation loss.}
Our deformation network $\dnet_i$ produces values $\sindicator_i$ that indicate a confidence with which a particular point belongs to geometric component $i$ (where components can be any of the teeth, or the gum).
These values are very important to ensure the locality of each deformation network: 
Only if component $i$ is sufficiently close to a given point should $\sindicator_i$ take a significant value and thus be able to contribute to the overall SDF value for that point. Any SDF contributions coming from components that are not closest to the point should be suppressed by multiplication with the blending weights.
Since the ground truth for segmentation is annotated on the surface, we will only actively supervise those sample points that reside on the surface with their ground truth labels. Since points that lie on the surface of one component are usually \emph{off} the surfaces of the other components, this strategy, in combination with the SDF loss on the global level (see below) is sufficient for making the $\sindicator_i$ values behave in the required way (see \cref{tab:ablation_accuracy} for an ablative evaluation).
We force each Deform-Net to solve a binary classification problem that can be formulated by point-wise binary cross-entropy (BCE) loss:

\begin{equation}
\loss^\segmentation_i=\sum_{\point \in \surface_i}\bce(\sindicator_i(\point),\lbl_i(\point)==i)
\end{equation}
where $\surface_i$ is the tooth or gum surface and $\lbl_i$ is the ground truth label for point $\point$.
The Deform-Net basically learns to classify surface locations as either belonging to component $i$, or \emph{not} belonging to component $i$.
Note that, in contrast to Mu~\etal~\cite{Mu2021ASDF}, who use one shared network to represent all shapes and multi-class segmentation labels, each of our submodules learns the deformation field and label of only one corresponding geometric component.

\paragraph{Deformation smoothness loss.} The deformation fields for all components should be rather smooth, because we expect the template shape learned by $\rnet_i$ to be close to the mean shape of component $i$. We thus constrain the deformation field $\dnet_i$ to be smooth, as in \cite{Deng2021DIF}:
\begin{equation}\label{eq:lsmooth}
\loss^\text{smooth}_i=\sum_{\point \in \spatial}\left\|\nabla \left(\point^\prime_i(\point)-\point\right)\right\|_2
\end{equation}
where  $\nabla(\point^\prime_i(\point) - \point)$ is the Jacobian of deformation offset with respect to the coordinates of $\point$ and $\spatial$ is the 3D spatial domain.

\paragraph{SDF loss.}
We do not use ground-truth SDF values for supervision
Instead we use a loss term from previous work \cite{Sitzmann2020Siren} that merely ensures that SDF values predicted by our model behave in a way that is consistent with the surface normals of the geometry and that satisfies the conditions that are generally expected from a signed distance field.
We use this loss term twice, namely on the component level and also on the global level:

\paragraph{Component-level SDF loss.}
For each component we minimize:

\begin{equation}\label{eq:sdflocal}
\begin{aligned}
\loss^{\text{SDF}}_i &=\sum_{\point \in \surface_i}\left|\model_i(\point)\right|+\left(1-\left\langle\nabla \model_i(\point), \bar{n}\right\rangle\right)\\
&+\sum_{p \in \Omega}\left|\left\|\nabla \model_i(\point)\right\|_{2}-1\right|+\sum_{\point \in \spatial \backslash \surface_i} \psi\left(\model_i(\point)\right)
\end{aligned}
\end{equation}
where $\psi(x)=\exp(-\alpha \cdot|x|)$ with $\alpha\gg1$. The first sum in this term encourages points on the surface $\surface_i$ to be mapped to SDF values close to 0, while the spatial gradient of the SDF value at those points should be directed parallel to the ground truth normal vectors $\bar{n}$. The last sum in the loss term forces \emph{non}-surface points to be mapped to SDF values that are \emph{different} from zero. The sum term in the middle forces the gradient of the signed distance field to be 1 almost everywhere.

\paragraph{Global-level SDF+Normal loss.} We use the same loss term also on the global level, to make sure that the result of blending all the geometric components together has the same desirable properties as each component individually:
\begin{equation}\label{eq:sdfglobal}
\begin{aligned}
\loss^{\text{SDF}} &=\sum_{\point \in \mathcal{S}}\left|f(\point)\right|+\left(1-\left\langle\nabla f(\point), \bar{n}\right\rangle\right)\\
&+\sum_{p \in \Omega}\left|\left\|\nabla f(\point)\right\|_{2}-1\right|+\sum_{\point \in \Omega \backslash \mathcal{S}} \psi\left(f(\point)\right)
\end{aligned}
\end{equation}

\paragraph{Normal consistency loss.}
The SDF loss (\cref{eq:sdflocal,eq:sdfglobal}) makes sure that our signed distance fields are consistent with the ground truth normal vectors for each instance. However, it is possible to satisfy the SDF loss while deforming points from different instances to different parts of the template shape. We need to discourage this, in order to establish consistent correspondences between the surface points of a particular instance and the surface points of the template shape learned by $\rnet$. As pointed out before \cite{Deng2021DIF}, this can be achieved
by forcing the normal directions at the points of the template surface to coincide with the normals at the corresponding points in each instance: 
\begin{equation}
\loss^\text{corres}_i=\sum_{\point \in \surface_{i}}\left(1-\left\langle\nabla \rnet_i\left(\point^\prime\right), \bar{n}\right\rangle\right)
\end{equation}
where $\nabla \rnet_i$ is the spatial gradient of the reference net, and $\bar{n}$ is the ground truth normal of the query point $\point$ on the original surface $\surface_i$.

\paragraph{SDF correction regularization loss.}
To make sure that SDF correction values $\dsd_i$ remain reasonably small and serve only to encode fine geometric details (as opposed to encoding the entire shape in general), we add the regularizer \cite{Deng2021DIF}
\begin{equation}
\loss^\text{correction}_i=\sum_{\point \in \spatial}|\dsd_i(\point)|
\end{equation}

\paragraph{Latent code regularization loss.} The latent codes are regularized by minimizing the term \cite{Park19DeepSDF}:
\begin{equation}
\loss^\text{latent}_i=\|\latent_i\|^2_2
\end{equation}

\noindent In summary, we optimize the following objective:
\begin{equation}
\begin{aligned}
\loss=\frac{1}{|\mathcal{I}|}\sum_{i\in\mathcal{I}}&(\lambda_1\loss^\centroid_i + \lambda_2\loss^\segmentation_i + \lambda_3\loss^\text{corres}_i+\lambda_4\loss^\text{smooth}_i\\
&+\lambda_5\loss^\text{latent}_i+\lambda_6\loss^\text{correction}_i+\lambda_7\loss^\text{SDF}_i)+\lambda_8\loss^{\text{SDF}}
\end{aligned}
\end{equation}
Here, $\mathcal{I}$ is the set of all component identities (i.e. all teeth and gum components) and the $\{\lambda_k\}$ are hyperparameters that control the weight of each term. Note that because dental patients may be missing one or more teeth, we dynamically set the component-level weights (i.e. $\lambda_1$ to $\lambda_7$) to zero in such cases, i.e. we do not supervise the submodules for components that are not present in a particular training scan.

\subsection{Data Preprocessing and Sampling}

In order to provide the training data for our morphable model, we first collect a set of dental scans (see \cref{fig:example_of_raw_data}).
Each of the scanned models was manually annotated, to segment them into gums and individual teeth, with each tooth being labelled by its identification number in the FDI World Dental Federation notation system (ISO-3950) \cite{ISO3950}.

Since the scans have been acquired using a variety of different devices, they are not aligned in a common coordinate system. We thus select one of the scans and normalize it to occupy the volume $[-1,1]^3$. Then we align all scans with this template, by aligning the centroids of corresponding teeth in a generalized Procrustes analysis (GPA) \cite{Gower1975GPA} that estimates the optimal scaling, rotation and translation for each instance, minimizing the sum of square distances. Note that the centroids of teeth are approximately on a plane which makes the GPA unstable. We handle this problem by checking the determinant of the rotation matrix and making sure that reflections are converted to rotations when necessary \cite{Arun1987GPA}.
Based on the aligned scans, we compute the average centroid for each tooth position.

In order to supervise our model, we need to sample the training data, which, for computational efficiency reasons, we do once before the start of training.
We follow the sampling strategy proposed in DIF-Net \cite{Deng2021DIF} with slight modifications to ensure that the distribution of the samples is adequate 
for capturing geometric details.  Specifically, we require most sample points to be on the surface of the geometry and we make sure that each tooth is sampled equally often, even though teeth differ in surface area.
For each dental scan, we sample 25,000 points on each tooth and 100,000 on the gum. All the surface sample points come with their labels and surface normals. We also sample 500,000 free space points uniformly from $[-1,1]^3$.

\section{Results}
\begin{figure}[t]
    \centering
    \includegraphics[width=0.95\columnwidth]{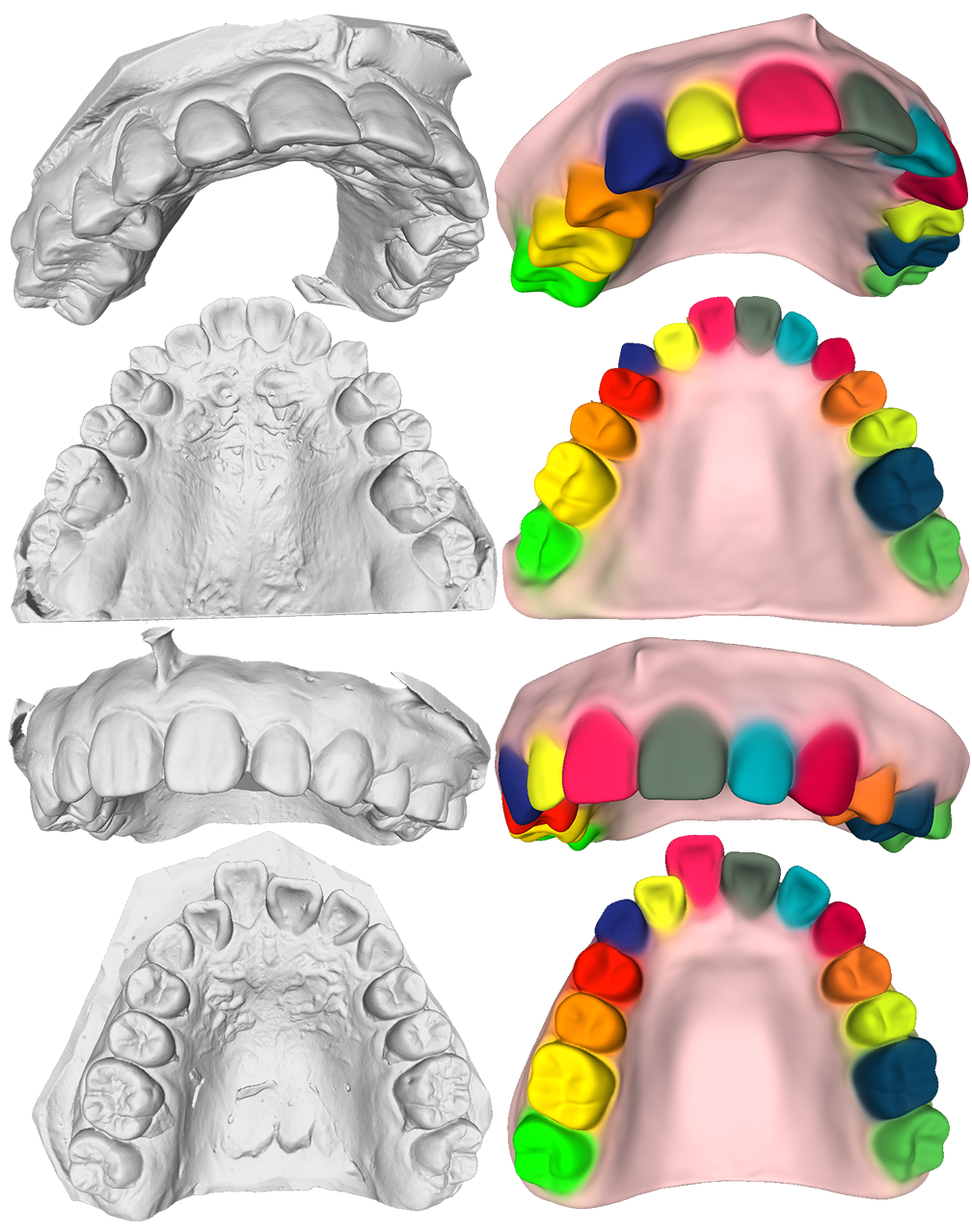}
    \caption{Reconstruction with teeth labelling. Left column: raw dental scan data; Right column: reconstruction results with teeth labelled by our method.}
    \label{fig:slabeling}
\end{figure}

\begin{figure}
    \centering
    \includegraphics[width=0.5\columnwidth]{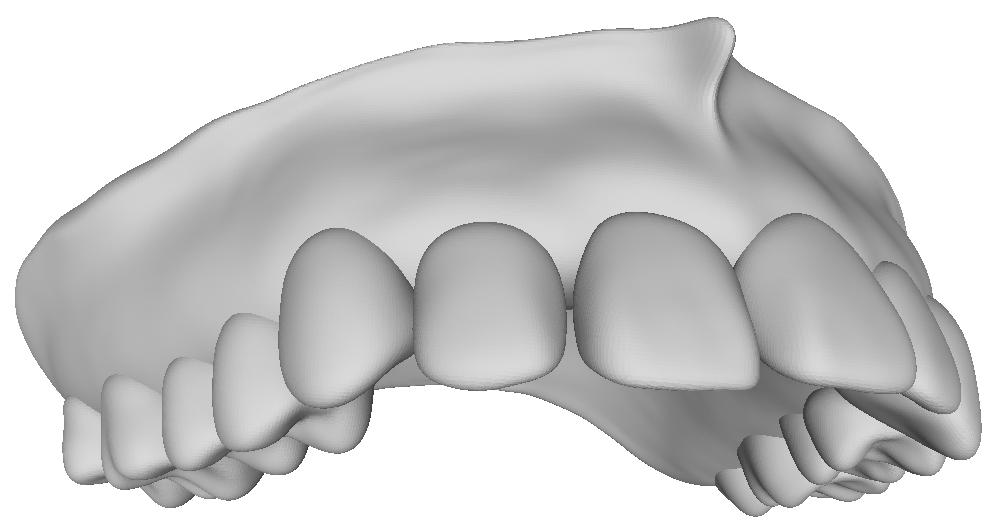}
  \caption{The template shapes learned by our submodules, combined into one overall geometry.}\label{fig:meanshape}
\end{figure}

\begin{figure*}[t]
    \centering
    \includegraphics[width=0.89\textwidth]{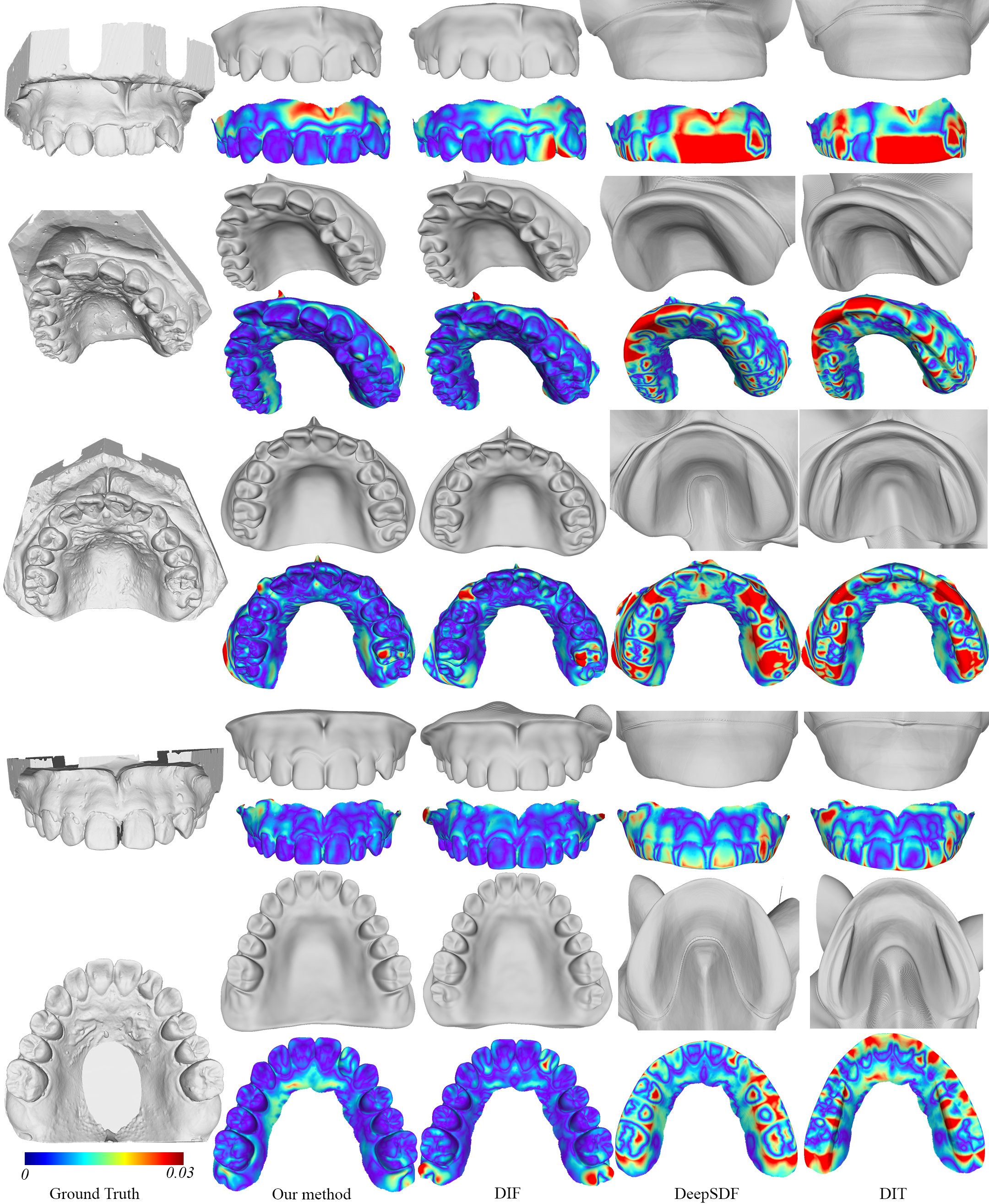}
    \caption{Comparison of reconstruction results with DIF \cite{Deng2021DIF}, DeepSDF \cite{Park19DeepSDF}, and DIT \cite{Zheng2021DIT}. Our method clearly outperforms DIT and DeepSDF. Furthermore, the error heat map shows that our method reconstructs the teeth region more accurately than DIF. Note that our method is the only one that offers independent control over each tooth and the gums, thus enabling interesting editing applications (see Sec.~\ref{sec:editing}).
    }
    \label{fig:reconstruction_gallery}
\end{figure*}

In this section, we evaluate our approach with regards to its reconstruction quality, its suitability for editing applications and the importance of its design choices.
We also compare it to a number of methods for implicit scene representations~\cite{Zheng2021DIT,Deng2021DIF,Park19DeepSDF}.
While each of these methods shares some of the features of our method, none of them decomposes the geometry into semantically delineated components in the way we do, and thus cannot provide separate control over the semantic components of the geometry (e.g. over individual teeth).
We refer the reader to the supplemental material for video results.

\subsection{Implementation details}\label{sec:implementation}
As described in \cref{sec:overview}, our model is trained on a dataset of dental scans.
This dataset contains 1077 maxilla geometries, about half of which are malaligned. We split them randomly into 1027 for training, and 50 for testing. By flipping left and right and interchanging labels, we augmented the training set to 2054 geometries.
\cref{fig:example_of_raw_data} shows some examples from the dataset.

For each geometric component (i.e. for each tooth type and for the gum), we use a latent code of length 10. We weigh our loss terms by $\lambda_1 = 1$, $\lambda_2 = 10^2$, $\lambda_3 = 10^2$, $\lambda_4 = 50$, $\lambda_5 = 10^6$, $\lambda_6 = 10^3$, $\lambda_7 = 1$, $\lambda_8 = 0.1$. Our model is trained on 2 NVIDIA Quadro RTX 8000 GPUs for 120 epochs, which takes about 36 hours.
In each iteration we sample 16384 points from 8 randomly selected ground truth scans. For all modules, the learning rate is $1 \cdot 10^ {-4}$, which we halve every 30 epochs.

\subsection{Reconstruction \& semantic labelling}\label{sec:reconstruction}

To reconstruct a given dental scan, we keep the weights of our trained model fixed and solve, similar to previous work \cite{Park19DeepSDF}, the following optimization problem:
\begin{equation}
\underset{\{\latent_i\}}{\arg \min } \left( \loss^\text{SDF}+\lambda^\text{latent}\sum_{i\in\mathcal{I}} \loss^\text{latent}_i \right)
\end{equation}
where $\mathcal{I}$ is the set of teeth indices that are present in the scan.
Note that, for a user of our method, merely providing a boolean vector that indicates the presence or absence of individual teeth is significantly easier than manually segmenting the individual teeth in the raw scan data.

In the first and second column of \cref{fig:reconstruction_gallery} we compare some ground truth dental scans to our reconstruction results. We observe that in general our reconstructions quite faithful, even in cases where teeth are severely misaligned, such as in the second row. In particular, our reconstructions exhibit clearly visible gum lines.

\cref{fig:meanshape} shows that our method also learns a meaningful template shape. It is via this template shape, that our method is not only able to \emph{reconstruct} a given dental scan, but also to \emph{label} it semantically, identifying each tooth in it (\cref{fig:slabeling}). The construction of our dataset still required this labelling to be done manually, but our method now provides a means of automating this task. 

\subsection{Editing applications}\label{sec:editing}

\begin{figure}[t]
    \centering
    \subcaptionbox[]{}{\includegraphics[width=0.32\columnwidth]{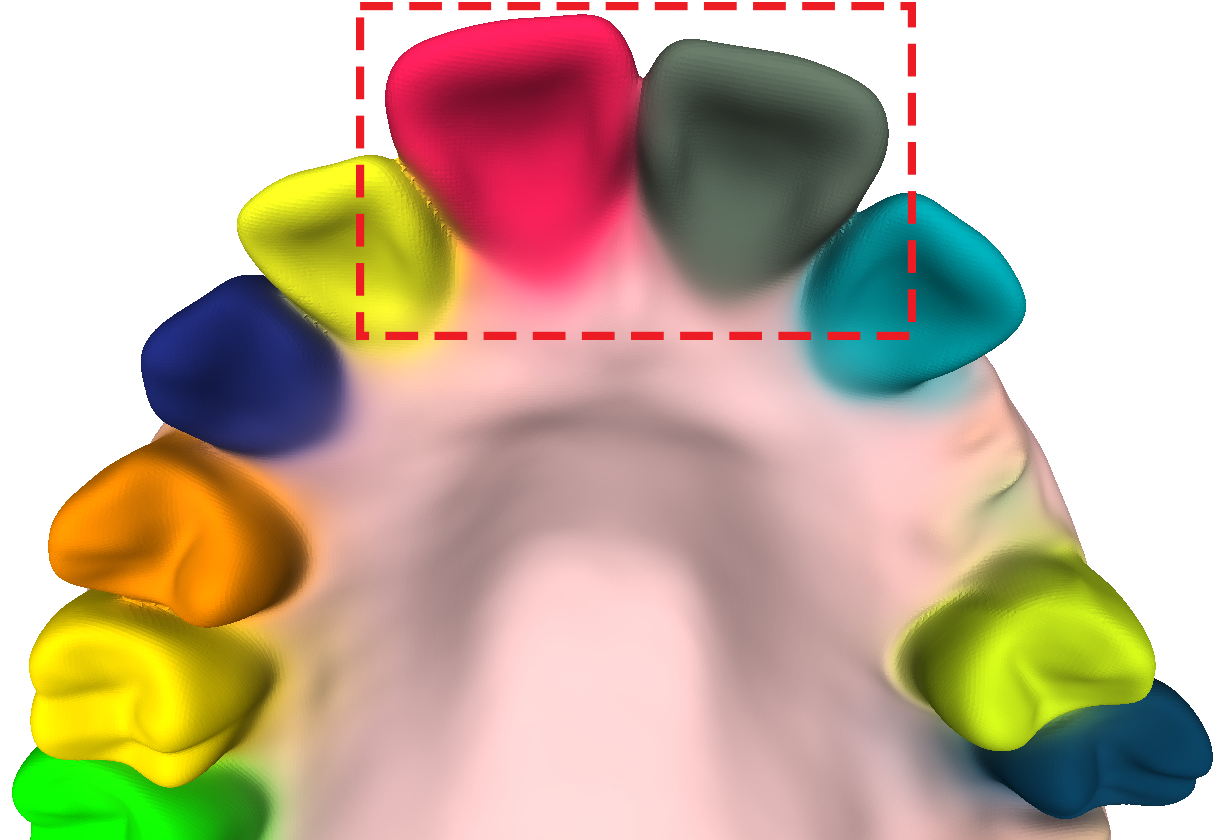}}
    \subcaptionbox[]{}{\includegraphics[width=0.32\columnwidth]{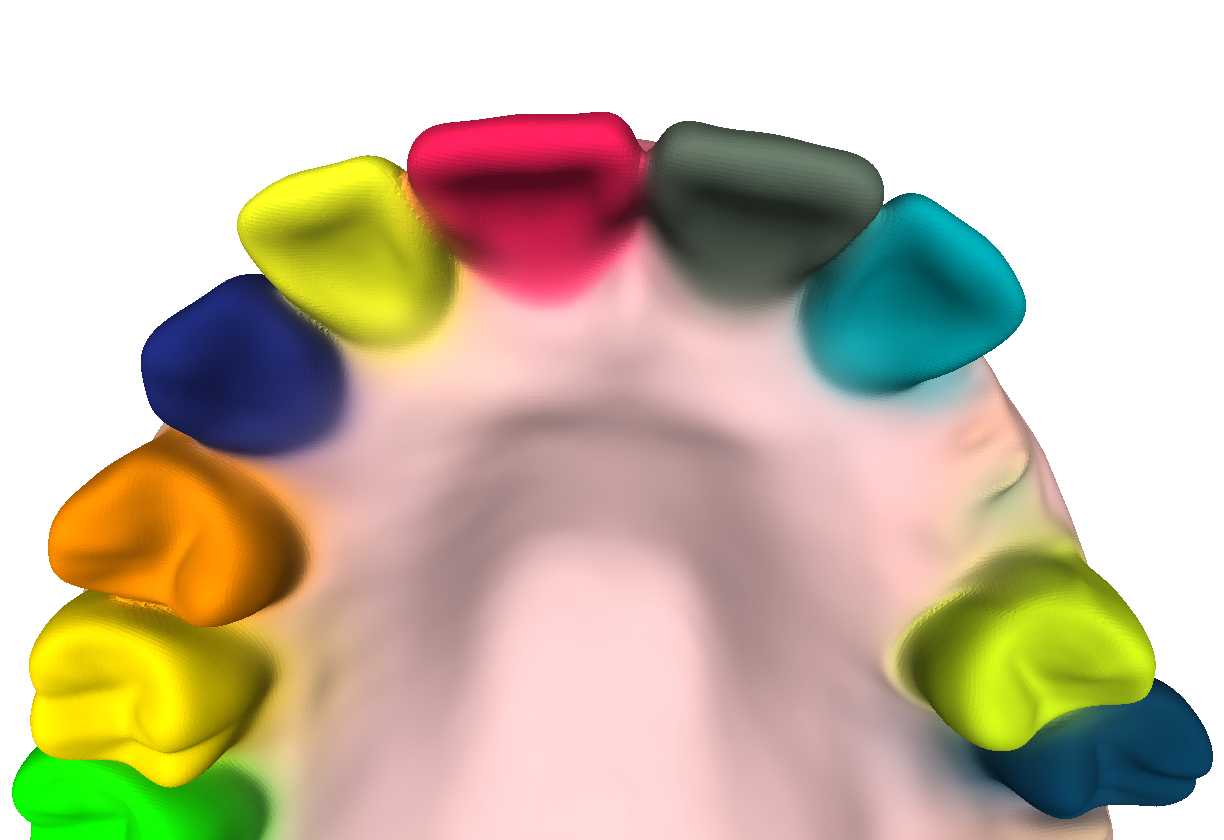}}
    \subcaptionbox[]{}{\includegraphics[width=0.32\columnwidth]{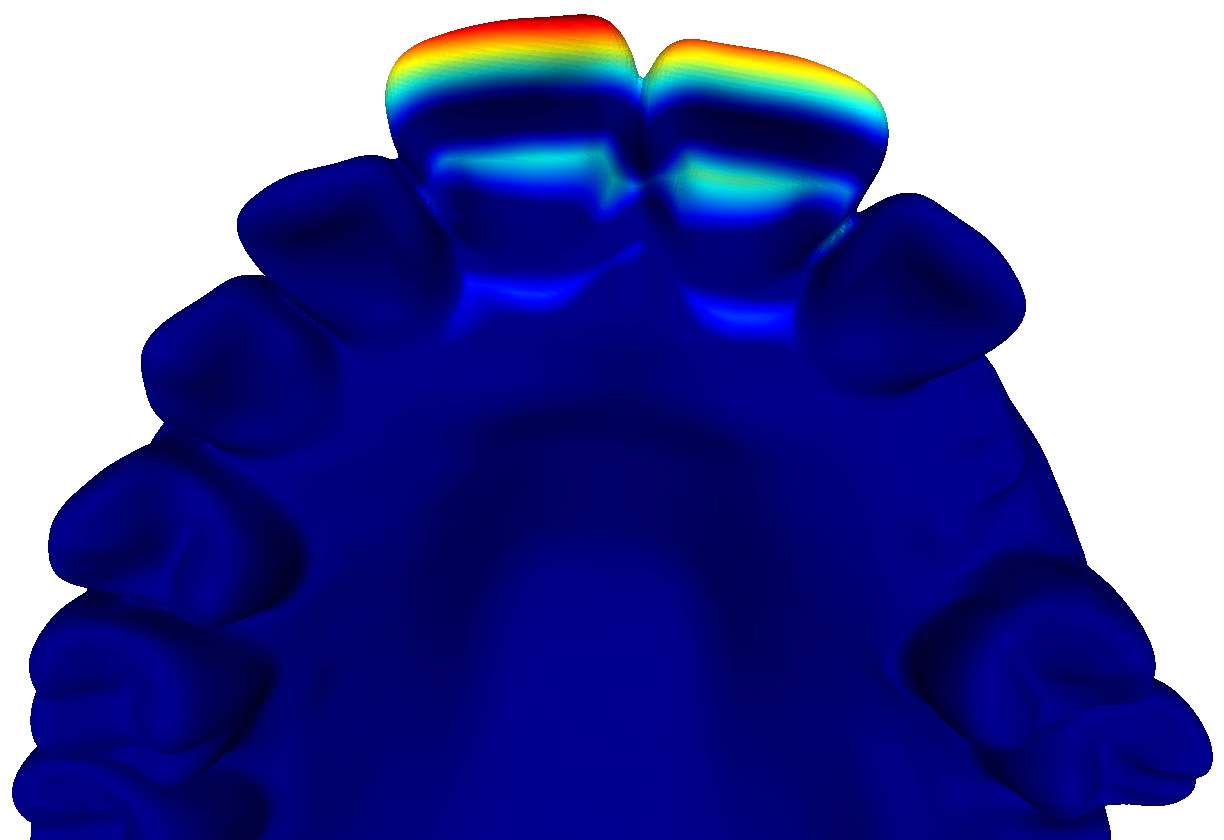}}\\
    \subcaptionbox[]{}{\includegraphics[width=0.32\columnwidth]{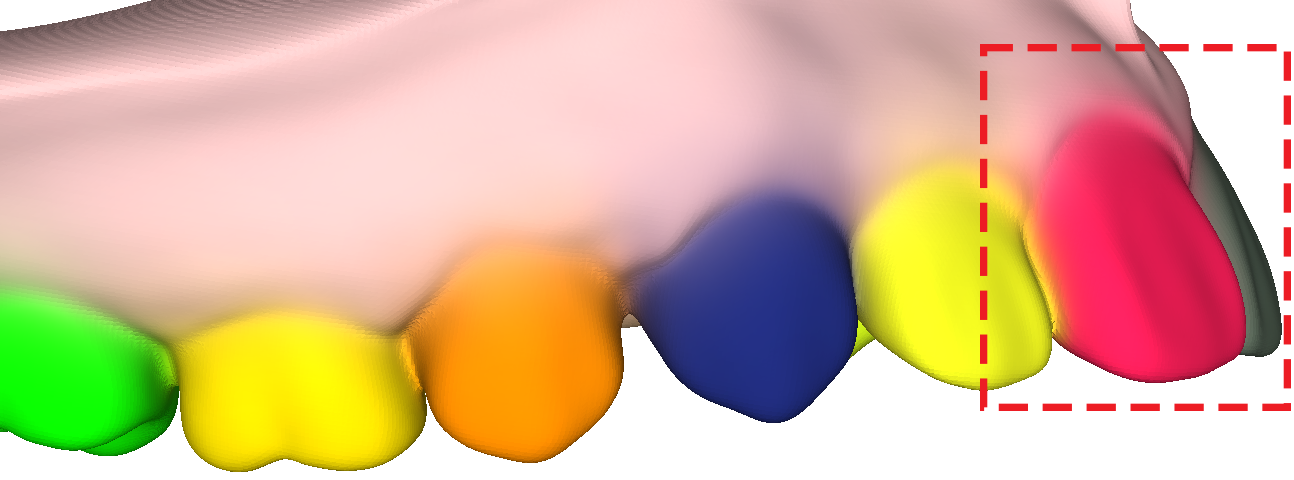}}
    \subcaptionbox[]{}{\includegraphics[width=0.32\columnwidth]{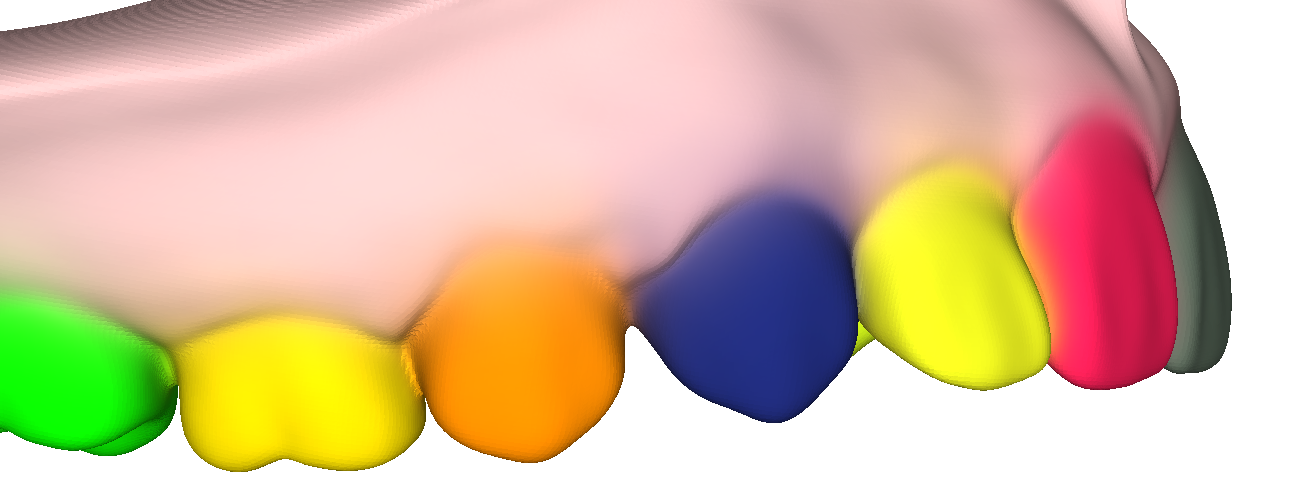}}
    \subcaptionbox[]{}{\includegraphics[width=0.32\columnwidth]{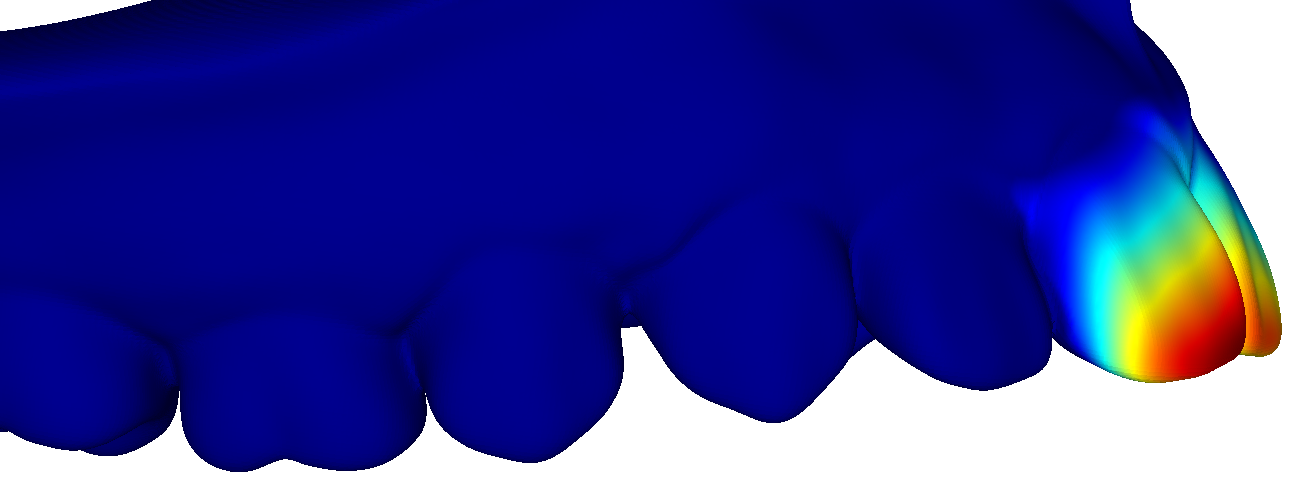}}
    \caption{Teeth replacement demonstration. (a) and (d) show two malaligned incisors from bottom and side view respectively (see dashed red). (b) and (e) show the result of replacing these two incisor teeth by some counterparts that are aligned better, while keeping all the other teeth unchanged.
     (c) and (f) encode the difference between before and after the edit.
    Note that the original model has no canine. Thus, we selected to process this example to show that we can reconstruct a model with originally missing teeth.}
    \label{fig:teeth_replacement}
\end{figure}

The major strength of our method is the fact that it decomposes teeth geometry into a number of semantically meaningful components that can be controlled individually. This allows us, for example, to edit a particular reconstruction result by replacing teeth that are mis-shaped or posed unaesthetically, by some more desirable counterparts (for example from a catalogue of aesthetically more pleasing teeth). \cref{fig:teeth_replacement} give an impression of this kind of editing application. Note that we only edited the incisor teeth, while all the other teeth remain unchanged. \cref{fig:teaser} gives another example (see dashed red).

\begin{figure*}[t]
    \centering
    \includegraphics[width=0.95\textwidth]{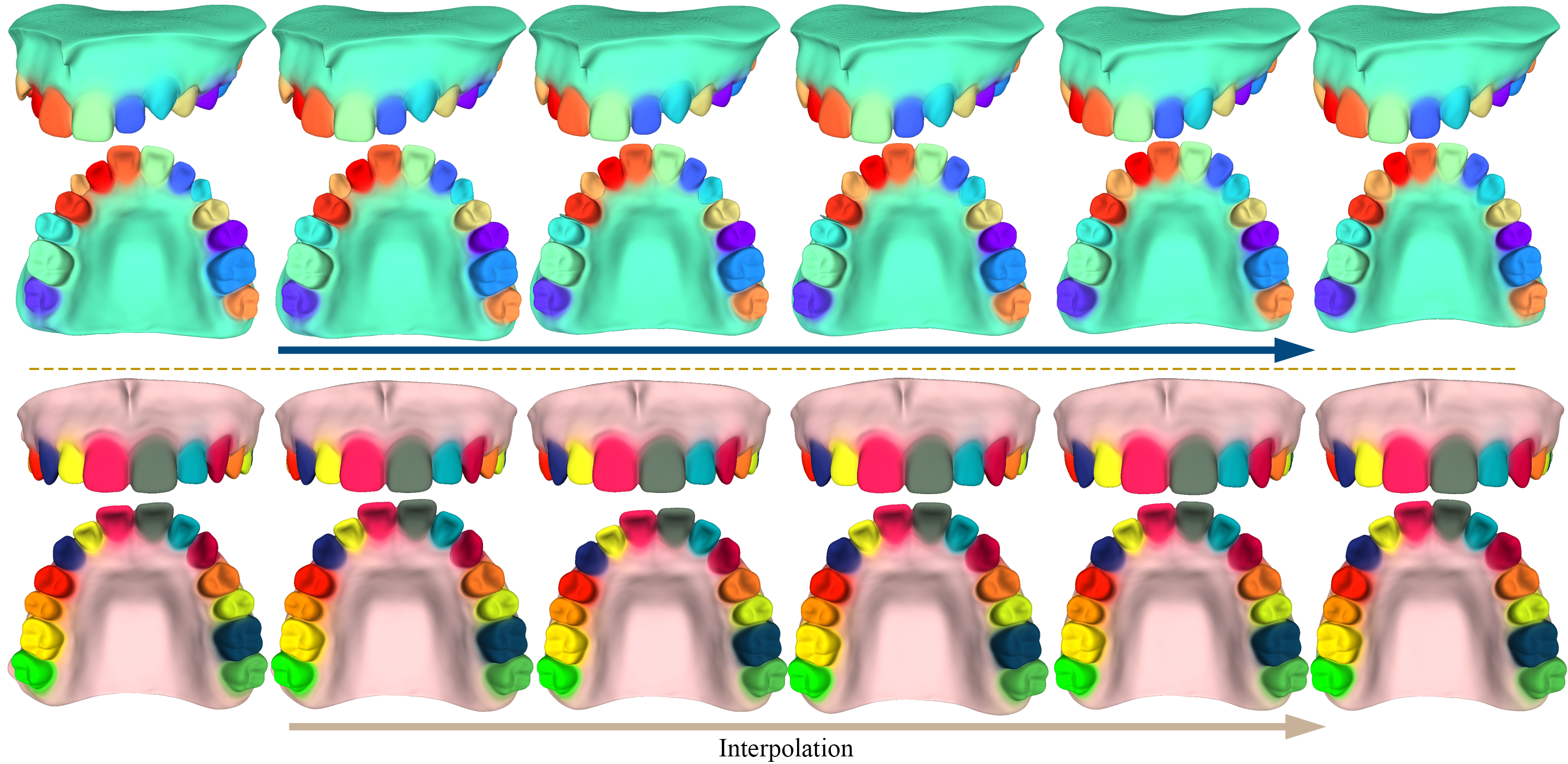}
    \caption{In each row we interpolate between the reconstruction of a pre-treatment scan (first column) and the reconstruction of a post-treatment scan (last column). 
    The arrows show the direction of interpolation. We can render plausible visualizations of orthodontic treatment plans in this way, which is best illustrated by our supplemental video results.
    }
    \label{fig:interpolation}
\end{figure*}

Especially in an orthodontic context, for example as part of a telehealth application, such editing could be used to visualize different treatment outcomes to a patient. Since orthodontic treatments such as correcting misaligned teeth can be a lengthy and continuous process, there may also be some merit in visualizing them as one continuous animation, which we illustrate in \cref{fig:interpolation} and in our supplemental video. Note that in those results we merely linearly interpolate latent codes from one teeth configuration to another. 
In an actual orthodontic use case, the animation would probably need to contain additional \enquote{keyframes}, based on orthodontic expert knowledge.

\subsection{Comparison to related works}

We compare our model to a number of implicit-based reconstruction methods, including the original DeepSDF \cite{Park19DeepSDF} and two more recent evolutions of it that both learn a template shape, DIT \cite{Zheng2021DIT} and DIF \cite{Deng2021DIF}. The latter is closest to our method, because it also uses Hyper-Nets to predict the weights of the deformation network. None of these methods, however use a component-wise representation and thus unlike our method they are unable to provide component-wise editing as we have shown in \cref{sec:editing}.

In \cref{fig:reconstruction_gallery} we compare our reconstruction results to those of the other methods. We observe that our reconstruction quality is clearly superior to that of DeepSDF or DIT. The comparison to DIF requires closer inspection: While at first sight DIF and our method seem to be on par, we consider the clearer gum line in our results as an advantage, given that this is also a clearly visible feature in the ground truth geometry. In the error maps we observe that while both methods struggle in similar areas of the gums, our method shows fewer and smaller erroneous regions on the teeth (see rows 1 and 3), especially when teeth are misaligned.

In \cref{tab:reconstruction_accuracy} we compare the methods numerically:
We extract meshes from  our reconstructions by marching cubes. 
After sampling a point cloud from such a mesh and from the ground truth mesh, we can compute symmetric Chamfer distance and an F-score based on that, as was done in previous work \cite{Yenamandra2021i3DMM} (after applying a threshold of 0.01 to the Chamfer distances). In all of our results, the width of the mean bounding box around the teeth geometry is approximately 2.

\cref{tab:reconstruction_accuracy} confirms that our method and DIF are very close in reconstruction quality and superior to the other methods. This is a very satisfying finding, given that our method is fulfilling the additional requirements of providing semantic labelling and allowing independent control over individual geometric components. DIF does not have these capabilities, but our method achieves them without compromising reconstruction quality at all.

\begin{table}[htbp]
\centering
\caption{Quantitative comparison with related works. The reconstruction accuracy is evaluated by the symmetric Chamfer distance (lower is better) and F-score (higher is better). Our overall reconstruction accuracy is on par with DIF. However, DIF does not enable the novel applications our method is capable of (Sec.~\ref{sec:editing}).
}\label{tab:reconstruction_accuracy}
\begin{tabular}{c|ll}
\toprule
Metrics     & Chamfer distance $\downarrow$  & F-score $\uparrow$  \\ \hline
\\[-1em]
DeepSDF     &  0.01497  & 42.132 \\ \hline
\\[-1em]
DIT         &  0.01353 &  47.668 \\ \hline
\\[-1em]
DIF         &  0.0058  &  \textbf{88.125} \\ \hline
\\[-1em]
Ours            &  \textbf{0.00552}   &   88.029  \\ \bottomrule
\end{tabular}
\end{table}

We also provide a comparison based on a publicly available 3D dental model dataset \cite{BenHamadou2022Benchmark}, which includes 595  dental models with ground truth labelling (after filtering out 2 duplicates and 3 cases with wisdom teeth). We split them into 545 for training (1090 after flipping data augmentation) and 50 for testing. The numerical results are shown in \cref{tab:reconstruction_accuracy_miccai}, confirming that our method achieves similar performance as DIF for the reconstruction task. All the methods yield better numerical results in \cref{tab:reconstruction_accuracy_miccai} than \cref{tab:reconstruction_accuracy} because the teeth in this dataset are more regular in shapes.

\begin{table}[htbp]
\centering
\caption{Quantitative comparison to related works on a publicly available dataset \cite{BenHamadou2022Benchmark}. The accuracy and F-scores are evaluated in the same way as \cref{tab:reconstruction_accuracy}. Similarly, our overall reconstruction accuracy is on par with DIF.}\label{tab:reconstruction_accuracy_miccai}
\begin{tabular}{c|ll}
\toprule
Metrics     & Chamfer distance $\downarrow$  & F-score $\uparrow$  \\ \hline
\\[-1em]
DeepSDF     &  0.01196  & 53.761 \\ \hline
\\[-1em]
DIT         &  0.01263 &  50.317 \\ \hline
\\[-1em]
DIF         &  0.00514  &  \textbf{92.622} \\ \hline
\\[-1em]
Ours            &  \textbf{0.00463}   &   92.182  \\ \bottomrule
\end{tabular}
\end{table}

\subsection{Ablation Study}
\begin{figure*}[t]
    \centering
    \includegraphics[width=0.9\textwidth]{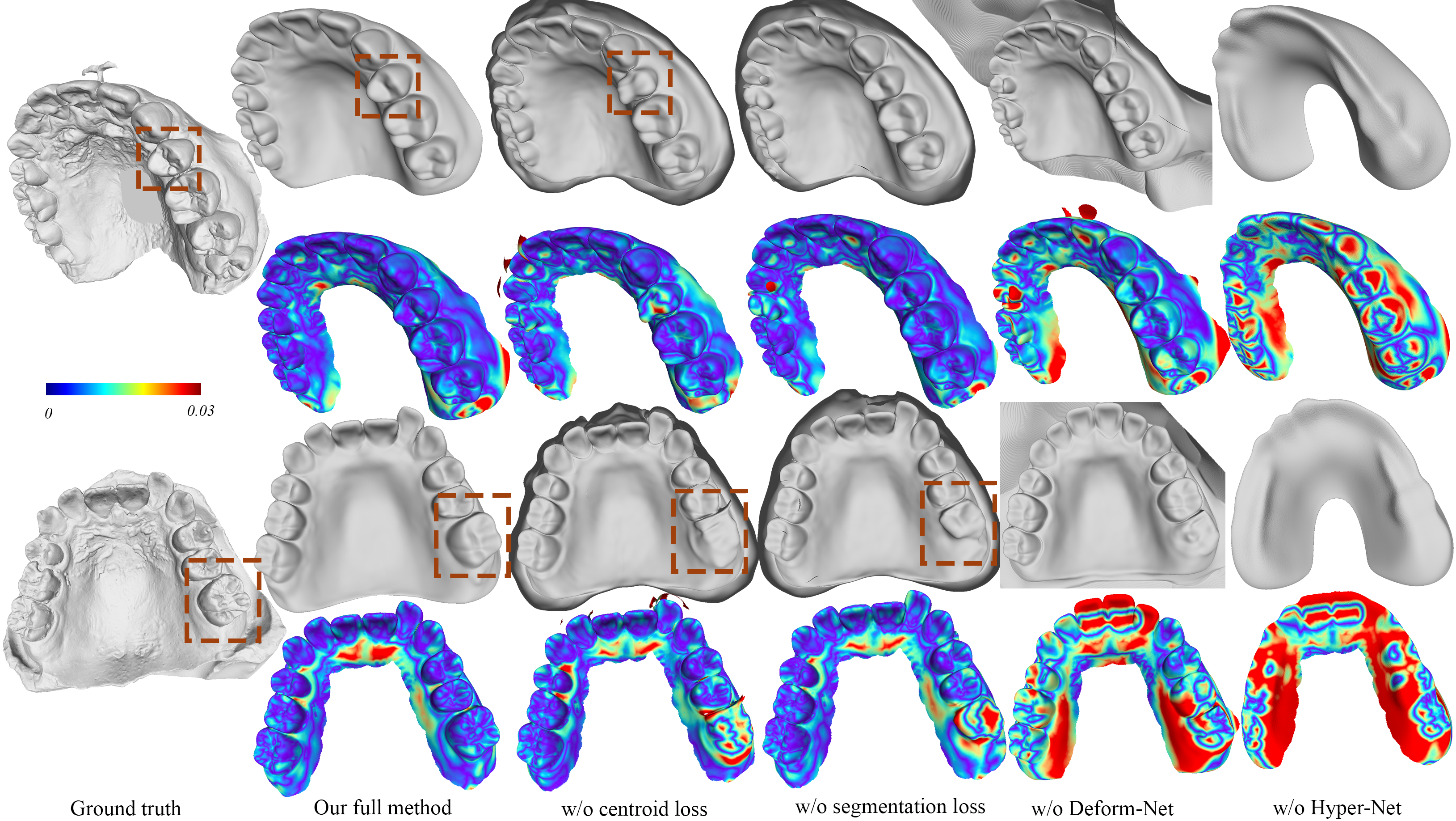}
    \caption{Similar to \cref{fig:reconstruction_gallery}, we visualize reconstruction results of our method and its ablated variants. As is confirmed by \cref{tab:ablation_accuracy}, our full method achieves the best results. Especially omitting our centroid loss or our segmentation loss can lead to severe artifacts (see dashed red).
    }
    \label{fig:ablation_gallery}
\end{figure*}

\begin{figure}[t]
    \centering
    \subcaptionbox[]{Our full method}{\includegraphics[width=0.48\columnwidth]{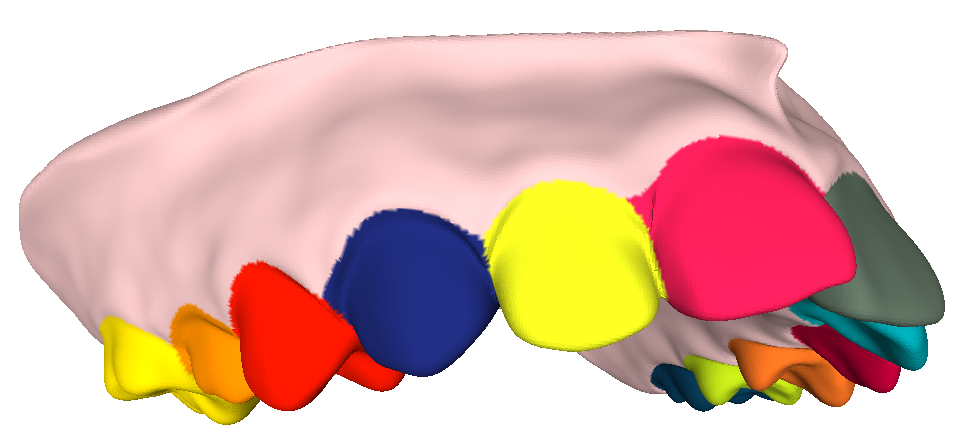}}
    \subcaptionbox[]{w/o Segmentation loss}{\includegraphics[width=0.48\columnwidth]{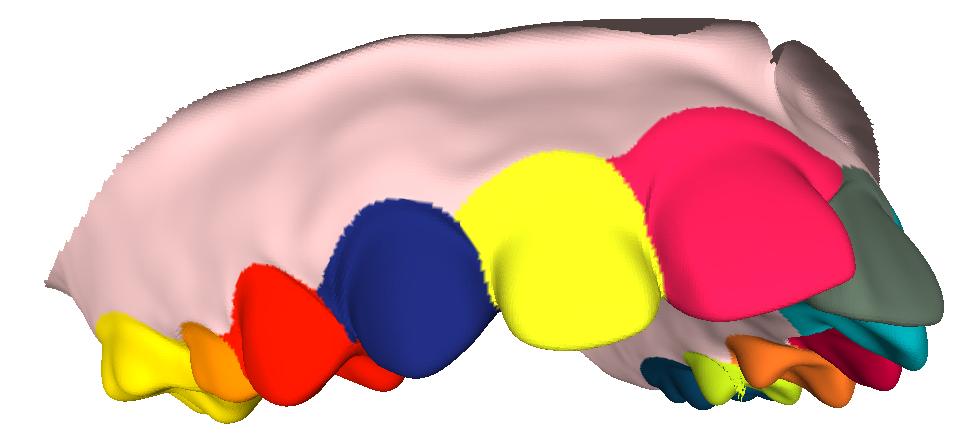}}
    \caption{Disabling our segmentation loss (b) clearly throws off the semantic labelling that our method (a) provides.
    }
    \label{fig:ablation_labelling}
\end{figure}

To evaluate our design choices, we have conducted an ablation study, comparing our full method to variants that lack our centroid loss or our segmentation loss (see \cref{sec:method}), as well as variants that omit the Deform-Net or train its weights directly, without obtaining them via a Hyper-Net.
\cref{tab:ablation_accuracy} shows that removing any of these design choices strongly deteriorates reconstruction accuracy. Our centroid and segmentation losses, which are absolutely vital for segmenting the geometry into separate semantic components also seem to improve reconstruction quality, as numbers consistently get worse when any of them is omitted.
The importance of these two losses, that we introduce in this work, is further highlighted by \cref{fig:ablation_labelling,fig:ablation_gallery}: \cref{fig:ablation_labelling} clearly shows that semantic labelling becomes very inaccurate when our segmentation loss is omitted. 
Note that the semantic labelling is not only a visualization but also an indicator of the locality of each individual component. When the semantic labelling is not accurate, the SDF contribution from each component is also not accurate.
\cref{fig:ablation_gallery} is consistent with \cref{tab:ablation_accuracy}: While all the design choices including the deformation field and the use of Hyper-Net have clear impact on the final result, omitting either centroid or segmentation loss can still significantly increase the error, for example as observed in the molar on the right in the second row (see dashed red).

\begin{table}[htbp]
\centering
\caption{Ablating the various design choices of our method. Our full method achieves the best results.}\label{tab:ablation_accuracy}
\begin{tabular}{l|cc}
\toprule
Metrics     & Chamfer distance $\downarrow$ & F-score $\uparrow$  \\ \hline
\\[-1em]
Our full method     &   \textbf{0.005522}      &   \textbf{88.029} \\ \hline
\\[-1em]
w/o centroid loss    &   0.006644     &  83.306  \\ \hline
\\[-1em]
w/o Segmentation loss   &  0.005985    &  86.706  \\ \hline
\\[-1em]
w/o Deform-Net         &   0.01377   &  51.818  \\ \hline
\\[-1em]
w/o Hyper-Net            &  0.01841  &  35.225        \\ \bottomrule
\end{tabular}
\end{table}
\section{Limitations \& Future Work}

\begin{figure}[t]
    \centering
    \subcaptionbox[]{}{\includegraphics[width=0.32\columnwidth]{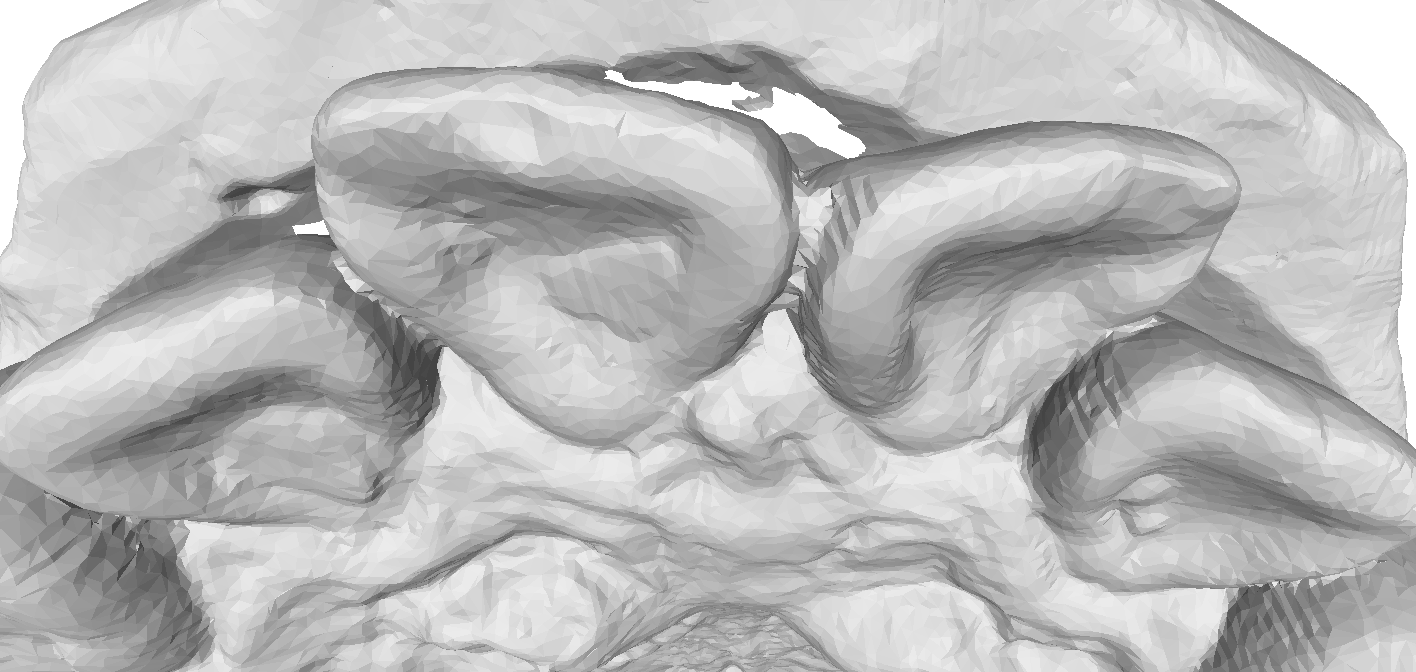}}
    \subcaptionbox[]{}{\includegraphics[width=0.32\columnwidth]{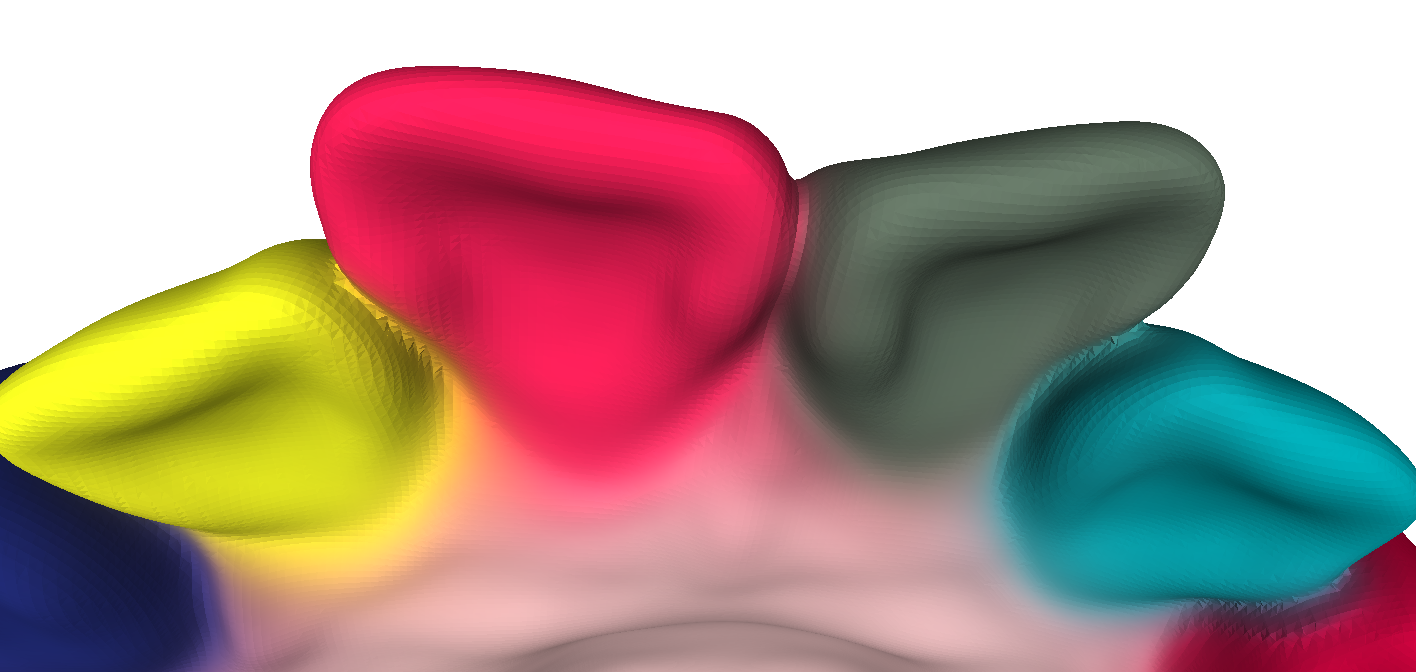}}
    \subcaptionbox[]{}{\includegraphics[width=0.32\columnwidth]{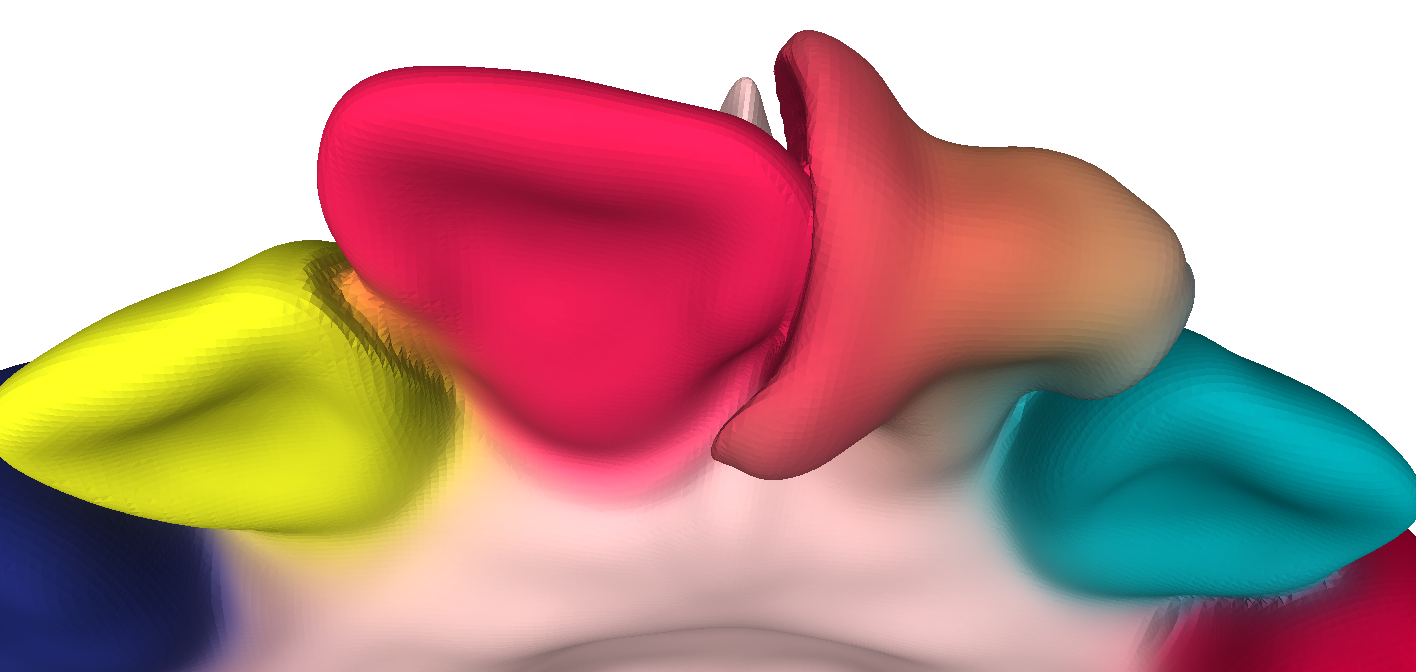}}
    \caption{ If a boolean vector indicating the presence/absence of individual teeth is provided, the ground truth geometry (a) can be faithfully reconstructed (b) by our method. If one gives an incorrect boolean vector, e.g. specifying that the right incisor (green in (b)) is missing, the resulting reconstruction (c) can have strong artifacts.
    }
    \label{fig:limitation}
\end{figure}

Even though the availability of a morphable teeth model with some control over individual teeth is a very useful contribution to the state of the art in this area, our model still has some limitations.
For example, at reconstruction time, we assume that the user provides a binary vector indicating the presence/absence of teeth in their respective positions. Without this information, missing teeth might lead to flawed results, as shown in \cref{fig:limitation}
While providing the binary vector is not too difficult for a user (such as a doctor or even a patient), it would be more satisfying to be able to infer it automatically.
To make the model usable in a broader range of use cases, it would have to include texture, which we have not addressed in this work. In addition, we completely omit the modelling of the tongue, which would be one major step on the way from a dental model to a full-fledged intra-oral model. We believe the modelling of the tongue to be a very difficult problem, because it is not even clear what data can be acquired for tongues.
A limitation much more easily overcome would be the extension of our model to the lower jaw, which does not require any changes to the method itself, but just training it on an additional dataset.

\section{Conclusion}

We have presented the first compositional, implicit neural representation for the modelling of teeth+gum geometry. Not only does our representation achieve state of the art quality in reconstruction (\cref{sec:reconstruction}), but also it decomposes the geometry into a number of semantically meaningful components, i.e. into individual teeth and the gum. The benefits of this decomposition are two-fold: First, it makes our reconstruction approach provide a semantic labelling for teeth geometry as an additional by-product (\cref{fig:slabeling}). Second, it shows interesting editing applications, such as the replacement of individual teeth by more aesthetically desirable alternatives (\cref{sec:editing}). Together with the fact that our model can be used for rendering smooth interpolations between different teeth states, it could be a valuable tool for the communication between orthodontists and their patients.

Beyond the concrete domain of teeth geometry, the contributions of this work lie in the way we achieve the local-based decomposition of complex geometry into smaller semantically meaningful components, which we hope can be beneficial in other domains as well.
We will release the pre-trained model and the inference code, which will make it the first publicly available teeth model of its kind.

\begin{acks}
This work is partially supported by the ERC Consolidator Grant 4DRepLy (770784), GRF grants (17210419 and 17212120) from the RGC of Hong Kong and by Seed Fund for Basic Research (202111159232) from Hong Kong. The authors would like to thank Yanhong Lin, Zhiming Cui, and Ayush Tewari for their instructive
suggestion and generous help on the project.
\end{acks}

\bibliographystyle{ACM-Reference-Format}
\bibliography{ref}


\end{document}